\documentclass{article} % For LaTeX2e
\usepackage[preprint]{colm2026_conference}

\usepackage{microtype}
\usepackage{hyperref}
\usepackage{url}
\usepackage{booktabs}

\usepackage{wrapfig}

\usepackage{xcolor}
\usepackage{stmaryrd} % \llbracket and \rrbracket
\usepackage{tikz}
\usetikzlibrary{arrows.meta,positioning,calc}

\usepackage{algorithm}
\usepackage{algpseudocode}

\usepackage{amsmath,amssymb}
\usepackage{amsthm}
\usepackage{mathtools}
\usepackage{bbm}
\usepackage{enumitem}
\usepackage{microtype}

\usepackage{caption}
\usepackage{lineno}

\definecolor{darkblue}{rgb}{0, 0, 0.5}
\hypersetup{colorlinks=true, citecolor=darkblue, linkcolor=darkblue, urlcolor=darkblue}

\title{Representational Homomorphism Predicts and Improves Compositional Generalization In Transformer Language Model}

\author{Zhiyu An, Wan Du\\
University of California, Merced\\
\texttt{\{zan7, wdu3\}@ucmerced.edu}
}

\begin{document}

\ifcolmsubmission
\linenumbers
\fi

\maketitle

\begin{abstract}
Compositional generalization—the ability to interpret novel combinations of familiar components—remains a persistent challenge for neural networks. 
Behavioral evaluations reveal \emph{when} models fail but offer limited insight into \emph{why} failures arise at the representational level. 
We introduce \textit{Homomorphism Error} (HE), a structural metric that measures the inconsistency between a set of established rules for which words combine to form new meaning (linguistic syntax) and model's learned rules for which hidden states combine to form new states (semantic syntax).
We formulate this inconsistency as deviations from approximate homomorphisms between the linguistic expression algebra and a model’s hidden-state space.
We designed experiments to test if i) HE predicts compositional generalization performance, and ii) will regularizing for low HE during training improve such performance.
To avoid the effect of data spoilage, we train small decoder-only Transformers from scratch using an adapted version of established dataset, SCAN, for testing compositional generalization.
Across controlled experiments, HE predicts out-of-distribution (OOD) compositional generalization under noise injection, achieving $R^2=0.73$ correlation between HE and OOD accuracy. 
Ablations show that model depth has minimal effect on either HE or OOD accuracy, training data coverage exhibits threshold effects, and randomly inserted noise tokens increase HE. 
Intervention experiment shows that HE-regularized training significantly reduces HE ($p=1.1\times10^{-4}$) and yields a statistically significant improvement in OOD accuracy ($p=0.023$). Together, these results indicate the potential of HE to be both a diagnostic and an actionable training signal for improving compositional generalization.
Code to reproduce our experiments is open-sourced at \url{https://github.com/ryeii/Representational-Homomorphism-for-Transformer-Language-Models}
\end{abstract}

\section{Introduction}

Human language understanding is characterized by systematic compositionality—the ability to combine familiar components in novel ways to understand expressions never encountered before~\cite{fodor1988connectionism}\cite{montague1970universal}. For instance, once a person learns the meaning of ``jump twice'' and ``turn'' they can immediately comprehend ``turn twice'' without explicit instruction. This compositional capacity enables humans to generalize from limited experience to an infinite space of possible expressions.

Despite remarkable progress in natural language processing, modern neural networks struggle with systematic compositional generalization \cite{dziri2023faith}. Empirical studies using benchmarks like SCAN~\cite{lake2018generalization}, COGS~\cite{kim2020cogs}, and CFQ~\cite{keysers2019measuring} have repeatedly demonstrated that while models achieve high accuracy on training distributions, they fail catastrophically when tested on novel combinations of familiar components. This limitation poses fundamental questions about whether neural architectures can truly capture the algebraic nature of human language understanding.

Current approaches to evaluating compositional generalization primarily rely on behavioral measures—comparing model outputs against expected results on held-out test sets. While such measures reveal \emph{when} models fail to generalize, they provide limited insight into \emph{why} these failures occur. Understanding the internal mechanisms that support or hinder compositional reasoning requires examining how models represent and manipulate compositional structure in their hidden layers, beyond surface-level performance metrics.

In this work, we introduce \emph{Homomorphism Error} (HE), a structural metric that quantifies how well neural network representations preserve compositional operations. Drawing inspiration from abstract algebra, we formalize compositionality as approximate homomorphisms between expression spaces and their representations. Low homomorphism error indicates that a model's internal representations respect compositional structure—that is, the representation of a composed expression can be systematically derived from the representations of its components. High homomorphism error suggests entangled or memorization-driven representations that fail to capture underlying compositional principles.

We study small transformer models on controlled SCAN-style tasks, aiming to isolate representational structure rather than claim immediate scalability to large pretrained LMs.
We evaluate HE on a customized SCAN-style synthetic dataset that allows systematic control over compositional structure, training data coverage, and noise levels, along with held-out test sets for Out-Of-Distribution (OOD) compositional generalization accuracy. Empirically, HE distinguishes when models learn compositionally structured representations versus when they rely on spurious correlations or memorization strategies. Finally, we move beyond correlational analysis by introducing \emph{HE-regularized training}---a simple auxiliary objective that explicitly encourages approximate homomorphisms in intermediate layers---and show that reducing HE during training yields consistent improvements in OOD compositional generalization under noise.
Our key contributions are:
\begin{itemize}[leftmargin=*]
\item We formalize compositionality as approximate homomorphism between syntactic and semantic algebras, and introduce homomorphism error as a task-independent metric that assesses compositional structure directly from model representations, complementing existing behavioral evaluation methods.

\item Through controlled experiments on SCAN-style synthetic compositional tasks, we demonstrate that homomorphism error serves as a reliable predictor of out-of-distribution generalization performance, achieving $R^2 = 0.73$ correlation in our noise injection studies.

\item We introduce HE-regularized training as an explicit intervention on representational structure, and show that enforcing low modifier HE \emph{causally} improves OOD compositional generalization ($p=0.023$) under noisy training conditions.
\end{itemize}

\section{Related Work}

\textbf{Compositional Generalization Benchmarks.}
The systematic evaluation of compositional generalization in neural networks began with Lake and Baroni's introduction of the SCAN dataset~\cite{lake2018generalization}. SCAN demonstrated that sequence-to-sequence models, while achieving high training accuracy, failed catastrophically when tested on systematic recombinations of known components. This work established the empirical foundation for studying the systematicity challenge first articulated by Fodor and Pylyshyn~\cite{fodor1988connectionism}.
Building on SCAN's foundation, subsequent benchmarks have explored different facets of compositional generalization. COGS~\cite{kim2020cogs} introduced semantic parsing challenges with natural language, finding that Transformers achieved near-perfect in-distribution accuracy (96-99\%) but much lower out-of-distribution performance (16-35\%). The grounded SCAN (gSCAN) benchmark~\cite{ruis2020benchmark} extended compositional evaluation to situated language understanding, where meaning depends on visual context. CFQ~\cite{keysers2019measuring} provided large-scale evaluation through systematical train-test splits that maximize compound divergence while minimizing atom divergence.

\textbf{Theoretical Frameworks.}
Hupkes et al.~\cite{hupkes2020compositionality} provided a comprehensive taxonomic framework, identifying five key aspects of compositionality: systematicity, productivity, substitutivity, localism, and overgeneralization. This theoretical foundation has guided much subsequent work in evaluating and understanding compositional behavior. Recent surveys~\cite{sinha2024survey} have further connected compositionality to broader questions of generalization and human-like reasoning in AI systems.

The field has developed several quantitative approaches to measuring compositional generalization. Keysers et al.~\cite{keysers2019measuring} introduced compound divergence as a metric for assessing train-test split difficulty, finding strong negative correlations between compound divergence and model accuracy. Other work has connected systematic generalization to information entropy~\cite{wold2025systematic}, showing that generalization scales with the distributional properties of compositional components in training data.

\textbf{Architectural Solutions.}
Various architectural innovations have been proposed to improve compositional generalization. Meta-learning approaches, particularly the MLC (Meta-Learning for Compositionality) framework~\cite{lake2023human}, have shown that neural networks can achieve human-like systematicity when optimized specifically for compositional skills. Neuro-symbolic approaches like the Compositional Program Generator~\cite{klinger2023compositional} achieve perfect performance on compositional benchmarks with dramatically improved sample efficiency.

For Transformer architectures, improvements have come through auxiliary training objectives~\cite{jiang2021inducing}, curriculum learning with dataset cartography~\cite{ince2023harnessing}, and architectural modifications such as increased depth~\cite{merrill2025little}. Graph-based semantic parsing frameworks have shown particular promise for structural generalization tasks~\cite{petit2023structural}.

\textbf{Internal Representation Analysis.}
Understanding the internal mechanisms underlying compositional behavior has been addressed through various probing methodologies~\cite{belinkov2021probing}. Work in mathematical reasoning has demonstrated that neural networks can learn compositionally structured representations that reflect sub-expression meanings~\cite{russin2021compositional}. Recent neuroscience-inspired work has shown evidence for compositional representations through algebraic operations on brain activity patterns~\cite{ferrante2025evidence}.

However, existing approaches to measuring compositionality have primarily focused on behavioral evaluation or task-specific probing. Our homomorphism error metric differs by providing a principled, architecture-agnostic measure of how well models preserve compositional structure in their internal representations, independent of surface-level task performance. We further show that this structural signal can be used not only diagnostically but also as a training-time regularizer that improves compositional generalization.

\section{Methods}

We begin by introducing a general homomorphism-based metric, Homomorphism Error (HE). 
We then instantiate this metric for compositional language tasks in two example forms, modifier (unary composition) and sequence (binary composition). 
Finally, we present a training objective that explicitly encourages low HE in learned representations.

\subsection{Preliminaries}

Formally, a \textbf{homomorphism} is a map between two algebraic structures of the same type (e.g., two groups, two fields, two vector spaces) that preserves the respective operations of these structures.

Let $f : (G, \ast) \to (F, \cdot)$ be such a map. Then, for all $(x, y) \in G$, $f(x \ast y) = f(x) \cdot f(y)$.

\subsection{Homomorphism Error as Structural Metric for Compositionality}\label{Sec: homomorphism error}

We begin by formalizing compositionality in terms of homomorphisms between
linguistic syntactic expressions and their semantic expressions.
Let $\mathcal{P}$ denote a finite set of primitives and $\circ$ a syntactic
composition operator defined by a grammar $G$.
Let $\mathcal{E}$ be the set of expressions generated from $\mathcal{P}$ using $\circ$.
Each expression $e \in \mathcal{E}$ has an associated semantic interpretation
$\llbracket e \rrbracket \in \mathcal{S}$, where $(\mathcal{S}, \bullet)$ is a semantic algebra
with composition operator $\bullet$.

\textbf{Compositionality as homomorphism.}
We say the mapping $\llbracket \cdot \rrbracket : \mathcal{E} \to \mathcal{S}$ is
\emph{compositional} if for all $e_1, e_2 \in \mathcal{E}$,
\begin{equation}
    \llbracket e_1 \circ e_2 \rrbracket \;=\;
    \llbracket e_1 \rrbracket \,\bullet\, \llbracket e_2 \rrbracket .
\end{equation}
That is, the meaning of a composed expression is given by the composition of the meanings of its parts.

\textbf{Approximate homomorphism in language models.}
Consider a language model $M_\theta$ with hidden representation function
$\Phi_\ell: \mathcal{E} \to \mathbb{R}^d$ at layer $\ell$.
We introduce an auxiliary learnable operator $\star : \mathbb{R}^d \times \mathbb{R}^d \to \mathbb{R}^d$
(e.g.\ linear map, bilinear map, or MLP) trained to approximate compositionality at the
representation level.
The \emph{Homomorphism Error (HE)} at layer $\ell$ is
\begin{equation}
    \mathrm{HE}_\ell \;=\;
    \mathbb{E}_{(e_1,e_2)\sim \mathcal{D}}
    \Big[ d\big( \Phi_\ell(e_1 \circ e_2), \;
    \Phi_\ell(e_1) \star_\ell \Phi_\ell(e_2) \big) \Big],
\end{equation}
where $\mathcal{D}$ is a distribution over expressions and $d$ is a distance
metric such as mean squared error (MSE).
In practice, we extract pairs of compossible expressions from the training dataset where $\star_\ell$ learns to predict the representation of a composed expression from its components.
We instantiate $\star_\ell$ with three operator families (linear, bilinear, MLP) and report the average error across them to avoid biasing the analysis toward a particular functional form of composition.

\begin{wrapfigure}{r}{0.45\textwidth}
\centering
\vspace{-1em}
\resizebox{0.35\columnwidth}{!}{%
\begin{tikzpicture}[
    node distance=1.6cm and 2.0cm,
    >=Latex,
    alg/.style={draw, rounded corners, inner sep=4pt, font=\footnotesize},
    lbl/.style={font=\scriptsize},
    dashedarr/.style={dashed, -{Latex[length=1.5mm]}}
]

% Top row: expressions and semantics
\node[alg] (EE) {$(\mathcal{E}\!\times\!\mathcal{E})$};
\node[alg, right=of EE] (E) {$(\mathcal{E},\,\circ)$};
\node[alg, above=1.4cm of E] (S) {$(\mathcal{S},\,\bullet)$};

% Bottom row: representations (aligned with expressions)
\node[alg, below=of EE] (RR) {$(\mathbb{R}^d\!\times\!\mathbb{R}^d)$};
\node[alg, below=of E]  (R)  {$(\mathbb{R}^d,\,\star)$};

% Solid arrows (square)
\draw[->] (EE) -- node[lbl, above]{\(\circ\)} (E);
\draw[->] (RR) -- node[lbl, above]{\(\star\)} (R);
\draw[->] (EE) -- node[lbl, left]{\(\Phi\times\Phi\)} (RR);
\draw[->] (E)  -- node[lbl, right]{\(\Phi\)} (R);

% Semantics arrow (dashed)
\draw[dashedarr] (E) -- node[lbl, right]{\(\llbracket\cdot\rrbracket\)} (S);

% Behavioral readout from R to S (dashed, curved)
\draw[dashedarr] (R.north east) .. controls ($(R.north east)+(0.9,0.9)$) and ($(S.south east)+(0.3,-0.2)$) .. node[lbl, right,pos=0.5]{\(I\)} (S.south east);

% Annotation
\node[align=center, lbl, below=0.4cm of R] (he) {%
Approx.\ commutativity: \\
\(\Phi(e_1\!\circ\! e_2)\;\approx\;\Phi(e_1)\star\Phi(e_2)\) \\
measured by \(\mathrm{HE}\)};

\end{tikzpicture}
}
\caption{Compositionality as approximate homomorphism.
Solid arrows form the representation-level square;
dashed arrows show semantic interpretation $\llbracket\cdot\rrbracket$ and an evaluation/readout $I$. Homomorphism Error (HE) quantifies how well $\Phi$ preserves composition.}
\label{fig:he-diagram}
\vspace{-2em}
\end{wrapfigure}

\textbf{Interpretation.}
Low $\mathrm{HE}$ indicates that internal representations are
\emph{approximately homomorphic} with respect to the task’s compositional
structure, whereas high $\mathrm{HE}$ suggests entangled or memorization-driven
representations.
Thus, $\mathrm{HE}$ measures the extent to which compositional structure is linearly or non-linearly decodable from hidden states, independent of task accuracy.

\subsection{Homomorphism Error Probe Design}

Building on the general definition of Homomorphism Error (HE) in Section \ref{Sec: homomorphism error}, we distinguish
two forms of compositionality present specifically in the above dataset, namely \textit{modifier HE} and \textit{sequence HE}, by the number of parameters (unary/binary) that the compositional operation requires.

\textbf{Modifier HE.}
Modifier homomorphism concerns unary composition $m(e)$, such as \texttt{twice} and \texttt{thrice}.
For a representation function $\Phi$ at model layer $\ell$, we define
\begin{equation}
    \mathrm{HE}^{\mathrm{mod}}_\ell
    \;=\;
    \mathbb{E}_{(m,e)\sim\mathcal{D}_{\mathrm{mod}}}
    \Big[ d\big(\Phi_\ell(m(e)), \; \star_\ell^m(\Phi_\ell(e)) \big) \Big],
\end{equation}
where $\star_\ell^m : \mathbb{R}^d \to \mathbb{R}^d$ is a learned operator
specific to modifier $m$, and $d$ is a distance metric such as MSE.
Low $\mathrm{HE}^{\mathrm{mod}}$ indicates that modifiers are represented as
structure-preserving transformations.

\textbf{Sequence HE.}
Sequence homomorphism concerns binary composition $e_1 \, c \, e_2$.
For representation function $\Phi_\ell$, we define
\begin{equation}
    \mathrm{HE}^{\mathrm{seq}}_\ell
    =
    \mathbb{E}_{(e_1, c, e_2)\sim\mathcal{D}_{\mathrm{seq}}}
    \Big[ d\big(\Phi_\ell(e_1 \, c \, e_2), 
        \star_\ell^c(\Phi_\ell(e_1), \Phi_\ell(e_2)) \big) \Big],
\end{equation}
where $\star_\ell^c : \mathbb{R}^d \times \mathbb{R}^d \to \mathbb{R}^d$
is a learned binary operator associated with connector $c$.
Low $\mathrm{HE}^{\mathrm{seq}}$ indicates that connectors are
represented as structure-preserving composition operators.

When calculating Modifier HE, we extract (primitive, modifier, combined) triples from the training dataset, where the combined representation is the mean pooling of consecutive primitive and modifier token representations.
For Sequence HE, we extract (part1, part2, combined) triples where parts are primitive-initiated segments and the combined representation is their average.
We use this representation extraction procedure consistently for both HE evaluation and HE-regularized training.
Noise tokens are not compossible with any other tokens and are not included for HE calculation, thus only the HEs of meaningful compositions are measured.

\subsection{HE-Regularized Training for Causal Compositionality}\label{sec:he-regularized-training}

While Homomorphism Error (HE) provides a diagnostic measure of compositional structure, a natural question is whether enforcing low HE during training can \emph{causally} improve compositional generalization, rather than merely correlating with it.
To address this question, we introduce \emph{HE-regularized training}, a simple auxiliary-objective framework that explicitly encourages internal representations to respect compositional structure.

\subsubsection{Training Objective}

We focus on \emph{modifier compositionality}, i.e.\ unary operations of the form $m(e)$, which our prior analysis shows to be particularly sensitive to noise and predictive of out-of-distribution (OOD) generalization.
Let $M_\theta$ be a language model with hidden representation function $\Phi_\ell$ at layer $\ell$.
Given a modifier $m$ and expression $e$, we define the modifier homomorphism constraint
\begin{equation}
\Phi_\ell(m(e)) \;\approx\; \star^m_\ell(\Phi_\ell(e)),
\end{equation}
where $\star^m_\ell$ is a learnable operator specific to modifier $m$.

To enforce this constraint during training, we augment the standard causal language modeling objective with an HE regularization term:
\begin{equation}
\mathcal{L}(\theta)
\;=\;
\mathcal{L}_{\mathrm{CE}}(\theta)
\;+\;
\lambda \cdot
\frac{1}{|\mathcal{L}|}
\sum_{\ell \in \mathcal{L}}
\mathrm{HE}^{\mathrm{mod}}_\ell ,
\end{equation}
where $\mathcal{L}_{\mathrm{CE}}$ is the standard cross-entropy loss, $\lambda > 0$ controls the strength of the compositional regularizer, $\mathcal{L}$ is a set of layers at which homomorphism is enforced, and $\mathrm{HE}^{\mathrm{mod}}_\ell$ is the modifier homomorphism error at layer $\ell$.
In practice, we regularize two intermediate layers (layers 2 and 4 in a 4-layer Transformer), which empirically balances expressivity and stability.

\subsection{Modifier HE Regularizer}

For each modifier $m$, we parameterize $\star^m_\ell$ as a small two-layer MLP operating on the concatenation of primitive and modifier representations:
\begin{equation}
\star^m_\ell([\mathbf{h}_e ; \mathbf{h}_m])
\;=\;
\mathrm{MLP}_m([\mathbf{h}_e ; \mathbf{h}_m]),
\end{equation}
where $\mathbf{h}_e = \Phi_\ell(e)$ and $\mathbf{h}_m = \Phi_\ell(m)$.
The target composed representation is defined as the mean-pooled representation of the composed span:
\begin{equation}
\mathbf{h}_{m(e)} \;=\; \tfrac{1}{2}(\mathbf{h}_e + \mathbf{h}_m),
\end{equation}
where representations are extracted by mean pooling over the corresponding token span.
The modifier HE at layer $\ell$ is then
\begin{equation}
    \mathrm{HE}^{\mathrm{mod}}_\ell
\;=\;
\mathbb{E}_{(m,e)\sim\mathcal{D}_{\mathrm{mod}}}
\big[
\lVert
\star^m_\ell(\Phi_\ell(e), \Phi_\ell(m))-
\Phi_\ell(m(e))
\rVert_2^2
\big].
\end{equation}

\subsection{Efficient HE-Regularized Training}

To make HE regularization computationally efficient, we pre-mine a fixed pool of modifier pairs $(e,m(e))$ from the training set.
At each optimization step, we sample a small auxiliary HE batch from this pool and compute the regularization loss using an additional forward pass.
This design keeps the computational overhead modest while providing a stable and strong learning signal.
We summarize the full training procedure in Appendix \ref{appendix: algorithm}.

\section{Experiments}\label{sec:experiments}

We consider two research questions: (i) diagnostic measurement of compositional structure via HE and (ii) causal intervention via HE-regularized training.

\textbf{RQ1 (Diagnostic):} How does homomorphism error change as we vary (a) model size, (b) training data coverage (sparsity), and (c) noise injection, and how do these changes relate to OOD compositional generalization?

\textbf{RQ2 (Causal):} Does explicitly regularizing homomorphism error during training reduce HE and improve OOD accuracy, beyond what is explained by correlation?

We first describe the dataset and HE probes, then specify experiment designs as targeted tests for RQ1 and RQ2, and finally present results that directly answer each question.

\subsection{Dataset Construction}

We design a controlled synthetic dataset inspired by SCAN-style tasks in order to
probe compositional generalization.
Let $\mathcal{P}$ denote a finite set of \emph{primitives} (e.g., \texttt{walk},
\texttt{jump}, \texttt{look}, \texttt{turn}) that each map to atomic output sequences over
a target alphabet $\Sigma$.
We further introduce a set $\mathcal{M}$ of \emph{modifiers} (e.g.,
\texttt{twice}, \texttt{thrice}) that act as unary operators
on primitives, and a set $\mathcal{C}$ of \emph{connectors}
(e.g., \texttt{then}) that define binary composition.

Formally, the grammar $G$ for input expressions is defined as
\[
e \;::=\; p \;\mid\; m(e) \;\mid\; e_1 \, c \, e_2,\;\;\text{where}\;\; p \in \mathcal{P}, m \in \mathcal{M}, c \in \mathcal{C}.
\]
The semantics $\llbracket e \rrbracket$ is an output sequence
in $\Sigma^\ast$ defined compositionally by rules such as
\[
\llbracket m(e) \rrbracket \;=\; f_m(\llbracket e \rrbracket),
\qquad
\llbracket e_1 \, c \, e_2 \rrbracket \;=\;
g_c(\llbracket e_1 \rrbracket, \llbracket e_2 \rrbracket),
\]
where $f_m$ and $g_c$ are deterministic rewriting functions.
For example,
\begin{align*}
&\llbracket \texttt{jump twice} \rrbracket = \llbracket \texttt{jump} \rrbracket
\,\llbracket \texttt{jump} \rrbracket, \\
&\llbracket \texttt{look then walk} \rrbracket
= \llbracket \texttt{look} \rrbracket \,\llbracket \texttt{walk} \rrbracket.
\end{align*}
To study the effect of different amount of noise in the training dataset, we introduce a finite set of \textit{noise tokens} denoted by $\mathcal{K}$ (e.g. \texttt{foo}, \texttt{bar}, \texttt{baz}).
When constructing noisy datasets, the noise tokens are inserted at random positions in the prompts, but not in the outputs.
For example,
\begin{equation*}
    \underbrace{\texttt{foo jump bar thrice then look baz}}_{\text{input}} \mapsto \underbrace{\texttt{jump jump jump look}}_{\text{output}}
\end{equation*}
This dataset construction allows us to systematically control the number of
primitives, modifiers, connectors, and noise tokens during training, and to evaluate
generalization to held-out combinations.

% \begin{figure}[t]
%   \centering
%   \includegraphics[width=\linewidth]{figures/HE_method\ (1).pdf}
%   \caption{Illustration of the HE measuring procedure.}
%   \label{fig:he-procedure}
% \end{figure}

\subsection{Experiment Designs}

We conduct four experiment blocks. The first three answer \textbf{RQ1} by perturbing model size, training coverage, and noise while tracking both OOD generalization and HE. The fourth answers \textbf{RQ2} by intervening on HE through training-time regularization.

All experiments use held-out OOD test sets composed of expressions containing $5$ to $12$ primitives. $200$ unique expressions are sampled from the space of primitives with each number of primitives. Layer-wise $\mathrm{HE}^{\mathrm{mod}}$ and $\mathrm{HE}^{\mathrm{seq}}$ are computed to measure internal compositional structure.

\textbf{Model architecture and training.}
All models are decoder-only transformers trained with a causal language modeling objective.
All experiments use a set of fixed hyperparameters: hidden dimension $d=128$, number of attention heads $h=4$, feedforward dimension $256$.
Inputs are tokenized and passed through learned embeddings with positional encodings.
The final hidden states are projected to the vocabulary space via a linear output layer. Models are trained using cross-entropy loss with teacher forcing, optimized with Adam ($\beta_1=0.9, \beta_2=0.98$), learning rate $10^{-4}$, and batch size $64$. Training runs for $50$ epochs with early stopping on validation loss. All experiments share the same optimization settings to isolate the effects of model depth, training sparsity, noise, and HE-regularization.

\textbf{RQ1a: Model size ablation.}
We vary the number of transformer layers $L \in \{1, 2, \dots, 10\}$.
For each configuration, we train models on a fixed dataset containing $2$ primitives with no noise, then evaluate OOD accuracy and both HEs.

\textbf{RQ1b: Training data sparsity.}
We construct datasets with expressions containing increasing numbers of primitives: $1,2,3,4$, always keeping $\mathrm{num\_noise}=0$.
For each sparsity level, models are trained with a fixed architecture ($L=4$ layers) and evaluated on the same OOD test sets. This isolates how \emph{structural coverage} in training affects HE and OOD accuracy.

\textbf{RQ1c: Noise injection.}
We construct training datasets with $2$ primitives and varying numbers of randomly inserted noise tokens $\mathrm{num\_noise} \in \{0,1,\dots,15\}$.
Models are trained with $4$ layers and evaluated on OOD sequences without noise. This tests whether spurious tokens disrupt internal compositional structure, as reflected in HE, and whether HE tracks the resulting OOD degradation.

\textbf{RQ2: HE-regularized training (causal intervention).}
We repeat the noise-injection setting but train a second set of models with the auxiliary HE regularizer from Section~\ref{sec:he-regularized-training}.
Comparing baseline vs.\ HE-regularized models under identical seeds and noise levels tests whether explicitly reducing HE during training reduces measured HE post hoc and improves OOD accuracy.

\textbf{Evaluation metrics.}
For each experiment, we report:
\begin{itemize}[leftmargin=*]
    \item \textbf{OOD accuracy:} average \% of correctly predicted output sequences on held-out test sets.
    \item \textbf{Modifier HE:} layer-wise MSE between representations of $m(e)$ and $\star_\ell^m(\Phi_\ell(e))$ across all modifiers. Values are final MSEs averaged across linear, bilinear, and MLP operators.
    \item \textbf{Sequence HE:} layer-wise MSE between representations of $e_1 \, c \, e_2$ and $\star_\ell^c(\Phi_\ell(e_1), \Phi_\ell(e_2))$ across all connectors. Values are final MSEs averaged across linear, bilinear, and MLP operators.
\end{itemize}

\textbf{Seed averaging and error bars.}
Each experiment is repeated with $5$ random seeds to control for stochasticity in dataset generation, model initialization, and training.
Reported metrics include mean and std across seeds, which are visualized as error bars in all plots.

\textbf{Computing Resources.}
All experiments (including $5$ random seeds per configuration across all ablations) were run on a single Apple M1 chip with $8$GB of memory. The full experiment suite completed in approximately $30$ minutes.

\subsection{Experiment Results}\label{sec:experiment-results}

\begin{figure}[t]
    \centering
    \includegraphics[width=.9\textwidth]{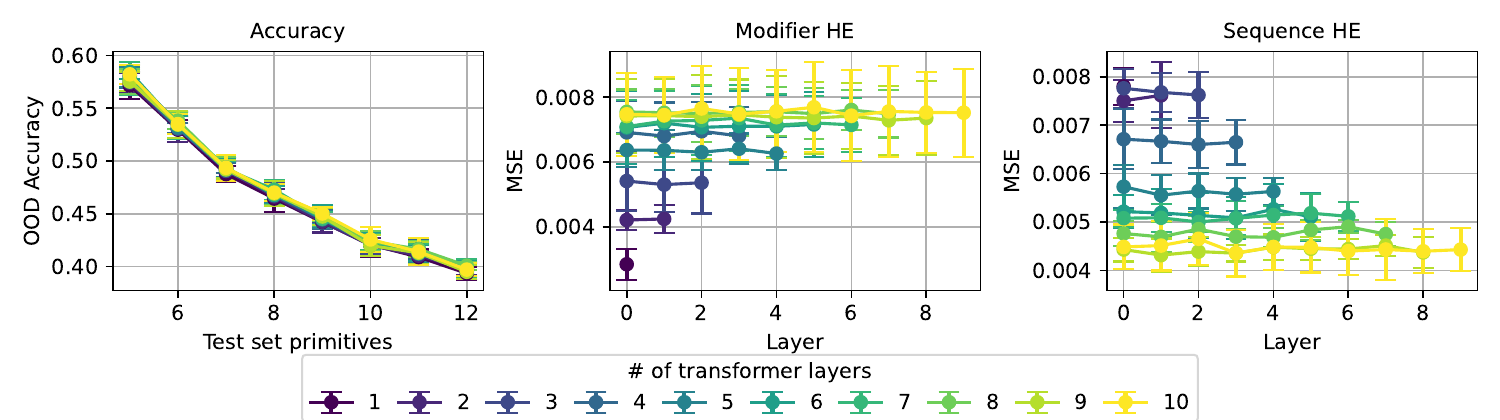}
    \caption{\textbf{RQ1a} Model size ablation. Lines represent number of transformer layers $L \in \{1, 2, \dots, 10\}$ in the language model.}
    \label{fig:model-size-ablation}
    \vspace{-1em}
\end{figure}

\begin{figure}[t]
    \centering
    \includegraphics[width=.9\textwidth]{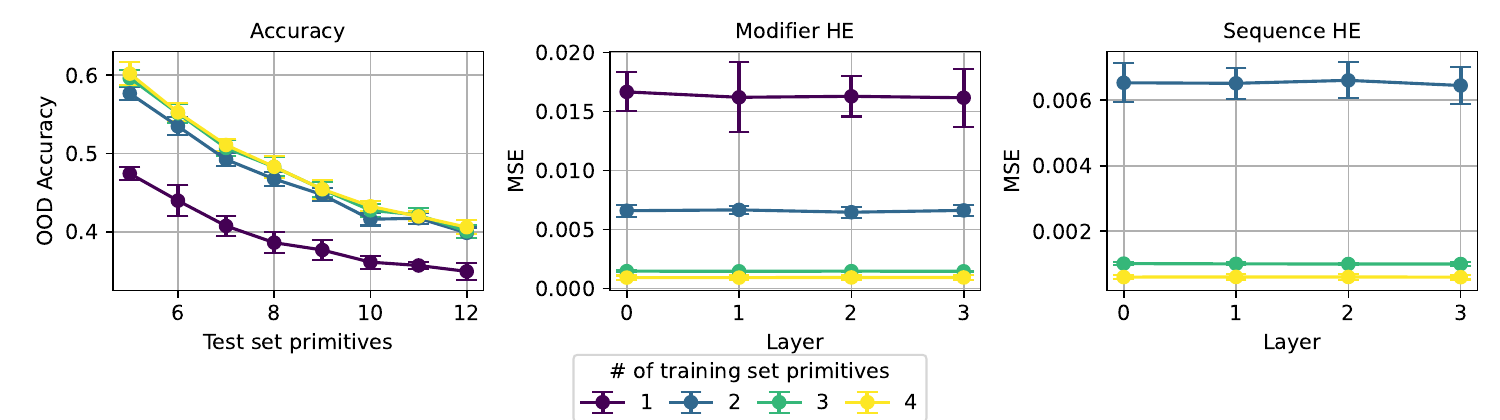}
    \caption{\textbf{RQ1b} Training data sparsity. Lines = maximum No. of primitives in the expressions presented in training set. $4$ means training set contains expressions with up to $4$ primitives.}
    \label{fig:sparsity-ablation}
    \vspace{-1em}
\end{figure}

\begin{figure}[t]
    \centering
    \includegraphics[width=.9\textwidth]{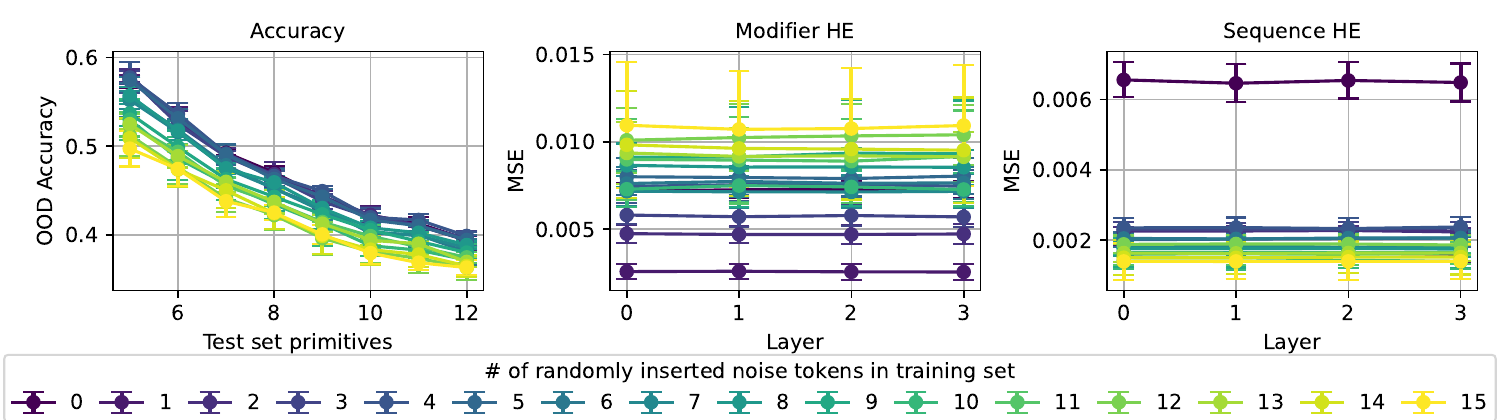}
    \caption{\textbf{RQ1c} Noise injection. Lines = number of noise tokens inserted in training data.}
    \label{fig:noise-ablation}
    \vspace{-1em}
\end{figure}

% Now we answer the research questions:

% \subsubsection*{RQ1: How does HE change with model size, sparsity, and noise, and how does that relate to OOD accuracy?}

\textbf{RQ1a (Model size): Model depth has minimal impact on OOD accuracy and HE.}
Figure~\ref{fig:model-size-ablation} shows that increasing model depth from 1 to 10 transformer layers yields negligible improvements in OOD accuracy (approximately 40--60\% across different test complexities). Both modifier and sequence homomorphism errors remain stable across model sizes, with variations on the order of $10^{-3}$. The modifier HE shows slight fluctuations between layers but no systematic trend, while sequence HE exhibits similarly minimal variation. This indicates that, in our controlled setting, representational capacity beyond a single layer provides little benefit for learning compositional structure.

\begin{wrapfigure}{r}{{0.5\textwidth}}
  \centering
  \vspace{-1em}
  \includegraphics[width=\linewidth]{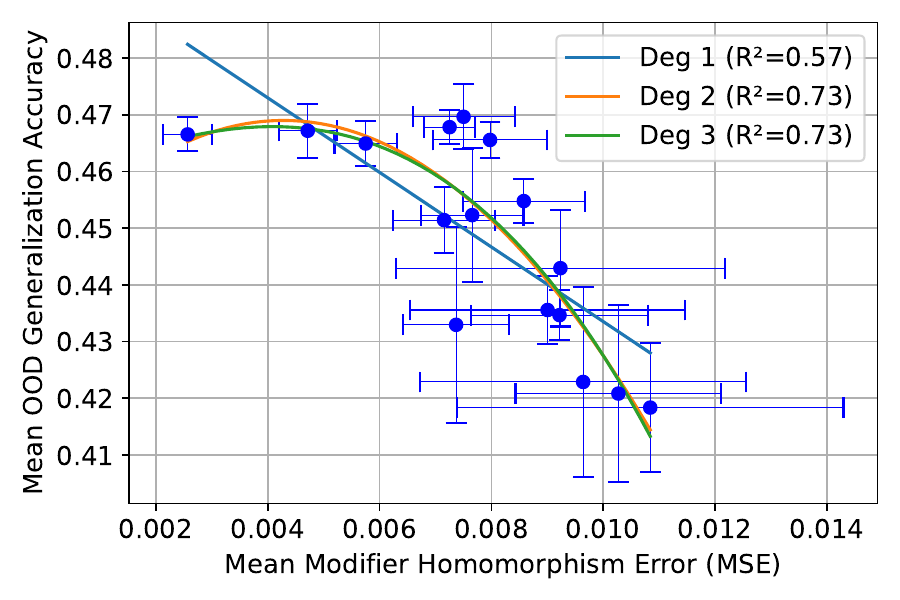}
  \vspace{-2em}
  \caption{Predictive relationship of HE and OOD accuracy under noise injection. 
  % Correlation between mean OOD generalization accuracy and mean modifier HE is shown. Polynomial regression with various degrees is conducted and $R^2$ is reported.
  }
  \label{fig:noise-acc-vs-he}
  \vspace{-1em}
\end{wrapfigure}

\textbf{RQ1b (Sparsity): Training coverage shows a threshold effect in HE and OOD generalization.}
Figure~\ref{fig:sparsity-ablation} reveals a sharp threshold effect. Models trained with only 1 primitive achieve lower OOD accuracy (35--47\%) and substantially higher modifier HE (0.015--0.018 MSE) compared to models trained with 2 or more primitives (40--60\% accuracy, 0.005--0.007 MSE modifier HE). This gap aligns with structural coverage: training sets with a single primitive cannot cover binary connectors, leaving models unable to learn sequence construction rules. Once training includes sufficient coverage (2+ primitives), adding more primitives yields little additional improvement in either OOD accuracy or HE.

\textbf{RQ1c (Noise): Noise injection systematically increases modifier HE and degrades OOD accuracy.}
Figure~\ref{fig:noise-ablation} shows that increasing the number of randomly inserted noise tokens produces a monotonic degradation in mean OOD accuracy (from approximately 47\% at 0 noise tokens to 42\% at 15 noise tokens). This is accompanied by a systematic increase in modifier HE (from approximately 0.002 MSE to 0.012 MSE), while sequence HE remains relatively stable across noise levels. This dissociation suggests that spurious tokens primarily disrupt unary (modifier) composition while leaving binary (sequence) composition comparatively intact.

\textbf{Predictive relationship under noise.}
The predictive power of our homomorphism error metric is illustrated in Figure~\ref{fig:noise-acc-vs-he}, which plots mean OOD accuracy against mean modifier HE across noise conditions. Polynomial regression yields $R^2 = 0.73$ for both quadratic and cubic fits, demonstrating that modifier HE strongly tracks OOD compositional generalization as noise increases.

\subsubsection*{RQ2: Does regularizing HE improve OOD accuracy?}

\begin{figure}
\centering
\begin{minipage}{.5\textwidth}
  \centering
    \includegraphics[width=.9\linewidth]{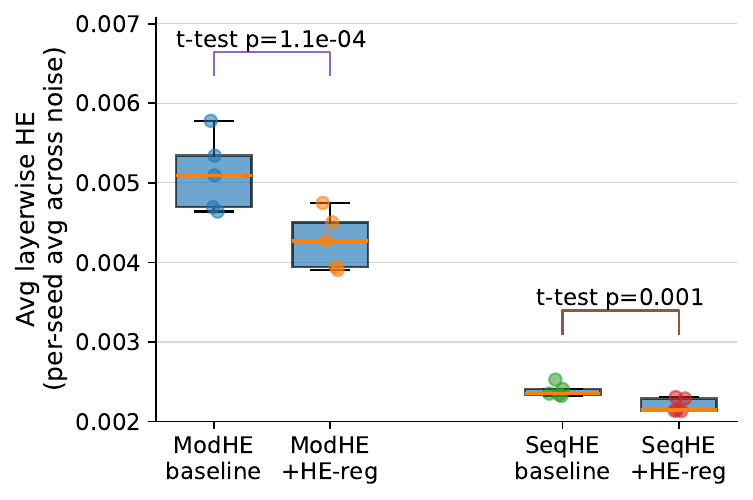}
    \captionsetup{width=.9\linewidth}
    \caption{
    Distribution of per-seed average layerwise HE across noise levels.
    % Box plots show modifier HE (left) and sequence HE (right) for baseline and HE-regularized models.
    % Each point corresponds to one random seed, averaged across noise conditions.
    % HE-regularized training significantly reduces modifier HE ($p=1.1\times10^{-4}$) and also reduces sequence HE ($p=0.001$), despite directly regularizing only unary composition.
    }
    \label{fig:he_reg_boxplot}
\end{minipage}%
\begin{minipage}{.5\textwidth}
  \centering
    \includegraphics[width=.9\linewidth]{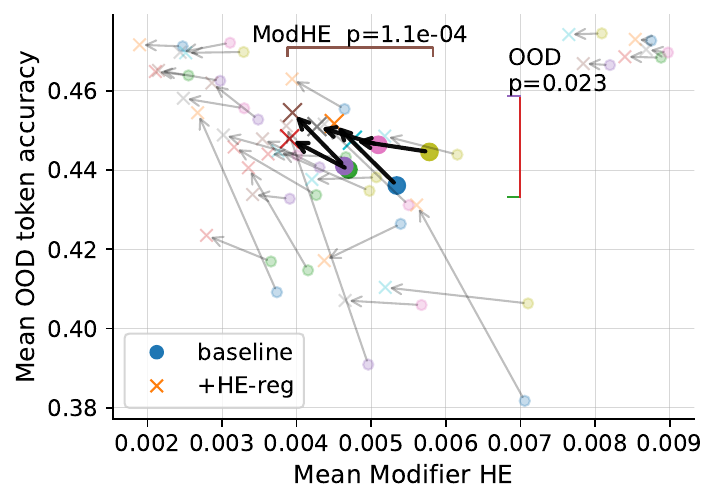}
    \captionsetup{width=.9\linewidth}
    \caption{
    \textbf{RQ2} Effect of adding modifier HE regularization on generalization.
    % Each point shows mean OOD token accuracy versus mean modifier HE.
    % Arrows connect baseline models to HE-regularized models for the same seed and noise level (faint), with bold arrows indicating per-seed averages.
    % HE-regularized training consistently reduces modifier HE ($p=1.1\times10^{-4}$) and yields a statistically significant improvement in OOD accuracy ($p=0.023$).
    }
    \label{fig:he_reg_scatter}
\end{minipage}
\vspace{-2em}
\end{figure}

\textbf{Effect on internal compositional structure.}
Figure~\ref{fig:he_reg_boxplot} compares per-seed average layerwise HE values between baseline and HE-regularized models (aggregated across noise levels). HE-regularized training yields a substantial and consistent reduction in modifier HE across seeds (paired t-test, $p=1.1\times10^{-4}$). Although the regularizer directly targets only unary modifier composition, sequence HE also decreases significantly (paired t-test, $p=0.001$), indicating that improving unary compositional structure can propagate to higher-level binary composition.

\textbf{Effect on compositional generalization.}
Figure~\ref{fig:he_reg_scatter} plots mean OOD accuracy against mean modifier HE for all runs. Arrows connect baseline models to their HE-regularized counterparts under the same seed and noise level. Across nearly all configurations, HE-regularization induces a consistent leftward shift (lower modifier HE) and an upward shift (higher OOD accuracy). Seed-level paired analysis confirms significantly higher OOD accuracy under HE-regularization (paired t-test, $p=0.023$). These results answer RQ2: enforcing approximate homomorphisms during training reduces measured HE and yields consistent improvements in OOD compositional generalization under noise.

% \textbf{RQ2 conclusion.}
% Together, these results provide causal evidence that homomorphism error is not merely diagnostic but functionally relevant to compositional generalization: explicitly enforcing low HE improves internal compositional structure and reliably translates into improved OOD accuracy under noisy supervision.

\section{Limitations}

Several limitations constrain the generalizability of our current findings. First, our experiments are conducted on synthetic data with precisely controlled compositional structure. While this enables rigorous analysis of the homomorphism error metric, real-world language presents additional complexities including semantic ambiguity, context dependence, and irregular constructions that may not conform to strict compositional principles.
Second, our evaluation focuses on relatively small Transformer models with controlled architectures and hyperparameters. The behavior of homomorphism error in large-scale pretrained language models remains an open question, particularly given evidence that scale can partially overcome compositional limitations~\cite{brown2020language}.

\section{Conclusion}

We introduced homomorphism error as a structural metric that formalizes compositionality as approximate homomorphisms between expression spaces and their representations.
Natural next steps including applying homomorphism error analysis to established compositional benchmarks using natural language, investigating how homomorphism error scales with model size and pretraining data, and exploring compositionally-aware architecture search and training procedures. 

\section*{Reproducibility Statement}

All code and data used to reproduce experiment results in this paper are open-sourced at 

\url{https://github.com/ryeii/Representational-Homomorphism-for-Transformer-Language-Models}

% \section*{LLM Usage Statement}

% LLM is used to provide editorial suggestions for this manuscript and code completion for the experiment implementation.

% \section*{Author Contributions}

% \section*{Acknowledgments}
% Use unnumbered first level headings for the acknowledgments. All

% \section*{Ethics Statement}
% Authors can add an optional ethics statement to the paper. 
% For papers that touch on ethical issues, this section will be evaluated as part of the review process. The ethics statement should come at the end of the paper. It does not count toward the page limit, but should not be more than 1 page. 

\bibliography{colm2026_conference}
\bibliographystyle{colm2026_conference}

\appendix
\section{HE-Regularized Training Algorithm}\label{appendix: algorithm}

We summarize the full HE-regularized training procedure for modifier HE below:

\begin{algorithm}[h]
\caption{HE-Regularized Training (Modifier Homomorphism)}
\label{alg:he-regularized-training}
\begin{algorithmic}[1]
\Require Training dataset $\mathcal{D}$, modifier pool $\mathcal{P}_{\mathrm{mod}}$, model $M_\theta$, layers $\mathcal{L}$, regularization weight $\lambda$
\For{each training step}
    \State Sample minibatch $\mathcal{B} \subset \mathcal{D}$
    \State Compute $\mathcal{L}_{\mathrm{CE}}$ on $\mathcal{B}$
    \State Sample modifier batch $\mathcal{B}_{\mathrm{mod}} \subset \mathcal{P}_{\mathrm{mod}}$
    \For{each layer $\ell \in \mathcal{L}$}
        \State Compute $\mathrm{HE}^{\mathrm{mod}}_\ell$ on $\mathcal{B}_{\mathrm{mod}}$
    \EndFor
    \State $\mathcal{L} \leftarrow \mathcal{L}_{\mathrm{CE}} + \lambda \cdot \frac{1}{|\mathcal{L}|} \sum_{\ell \in \mathcal{L}} \mathrm{HE}^{\mathrm{mod}}_\ell$
    \State Update $\theta$ via gradient descent on $\mathcal{L}$
\EndFor
\end{algorithmic}
\end{algorithm}

\end{document}